\documentclass{article}

\usepackage[preprint]{neurips_2022}

\usepackage[utf8]{inputenc} % allow utf-8 input
\usepackage[T1]{fontenc}    % use 8-bit T1 fonts
\usepackage{hyperref}       % hyperlinks
\usepackage{url}            % simple URL typesetting
\usepackage{booktabs}       % professional-quality tables
\usepackage{amsfonts}       % blackboard math symbols
\usepackage{nicefrac}       % compact symbols for 1/2, etc.
\usepackage{microtype}      % microtypography
\usepackage{xcolor}         % colors

\usepackage{epsfig}
\usepackage{graphicx}
\usepackage{bm}
\usepackage{multirow}
\usepackage{natbib}
\setcitestyle{numbers,square}
\usepackage{changes}
\usepackage{wrapfig}
\usepackage{hyperref}
\usepackage{algorithmic}
\usepackage[ruled]{algorithm2e}

\title{Temporally Resolution Decrement: Utilizing the Shape Consistency for Higher Computational Efficiency}

\author{%
  Tianshu Xie \\
  {\small University of Electronic Science }\\
  {\small and Technology of China}\\
  \texttt{tianshuxie@std.uestc.edu.cn} \\
  \And
  Xuan Cheng\thanks{Equally  contributed.} \\
  {\small University of Electronic Science }\\
  {\small and Technology of China}\\
  \texttt{cs\_xuancheng@std.uestc.edu.cn} \\
  \And
  Xiaomin Wang \\
  {\small University of Electronic Science }\\
  {\small and Technology of China}\\
  \texttt{xmwang@uestc.edu.cn} \\
  \And
  Minghui Liu \\
  {\small University of Electronic Science }\\
  {\small and Technology of China}\\
  \texttt{minghuiliuuestc@163.com} \\
  \And
  Jiali Deng \\
  {\small University of Electronic Science }\\
  {\small and Technology of China}\\
  \texttt{julia\_d@163.com} \\
  \And
  Ming Liu\thanks{Ming Liu is the corresponding author.}  \\
  {\small University of Electronic Science }\\
  {\small and Technology of China}\\
  \texttt{csmliu@uestc.edu.cn} \\
}
\begin{document}

\maketitle

\begin{abstract}
Image resolution that has close relations with accuracy and computational cost plays a pivotal role in network training. In this paper, we observe that the reduced image retains relatively complete shape semantics but loses extensive texture information. Inspired by the consistency of the shape semantics as well as the fragility of the texture information, we propose a novel training strategy named Temporally Resolution Decrement. Wherein, we randomly reduce the training images to a smaller resolution in the time domain. During the alternate training with the reduced images and the original images, the unstable texture information in the images results in a weaker correlation between the texture-related patterns and the correct label, naturally enforcing the model to rely more on shape properties that are robust and conform to the human decision rule. Surprisingly, our approach greatly improves both the training and inference efficiency of convolutional neural networks. On ImageNet classification, using only 33\% calculation quantity (randomly reducing the training image to 112$\times$112 within 90\% epochs) can still improve ResNet-50 from 76.32\% to 77.71\%. Superimposed with the strong training procedure of ResNet-50 on ImageNet, our method achieves 80.42\% top-1 accuracy with saving 37.5\% calculation overhead. To the best of our knowledge this is the highest ImageNet single-crop accuracy on ResNet-50 under 224$\times$224 without extra data or distillation.
\end{abstract}

\section{Introduction}

Looking back on the last decade, convolutional neural networks (CNNs) have enabled tremendous leaps of progress in computer vision, as reflected by the super-human-level performance on various benchmarks, such as ILSVRC’2012 challenge~\cite{krizhevsky2012imagenet} and the COCO detection competition~\cite{2014Microsoft}. However, the transformational advances usually come at complex network design with increasing depth or width, which would translate into high computation cost and memory footprint. In this regard, how to maintain or even improve CNNs’ performance while reducing the computational cost is undoubtedly worthy research, but it puts forward higher requirements for our understanding to CNNs. 

\begin{figure*}[t]
	\centering
	\hspace{0.01\linewidth}%
	\begin{minipage}[c]{0.43\textwidth} 
		\centering 
		\includegraphics[width=\linewidth]{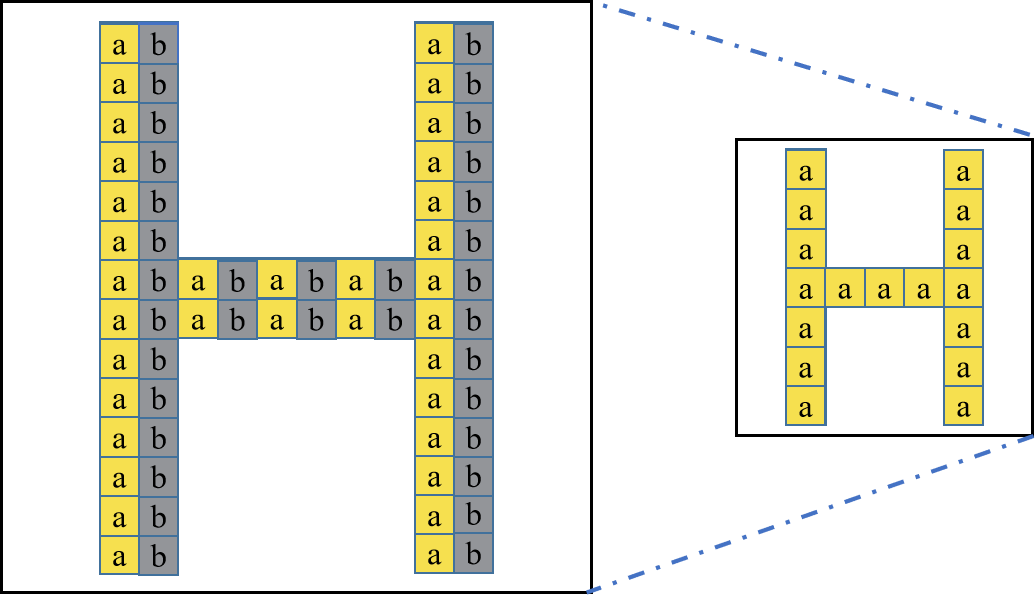} 
	\end{minipage}
	\hfill 
	\begin{minipage}[c]{0.43\textwidth} 
		\centering 
		\includegraphics[width=\linewidth]{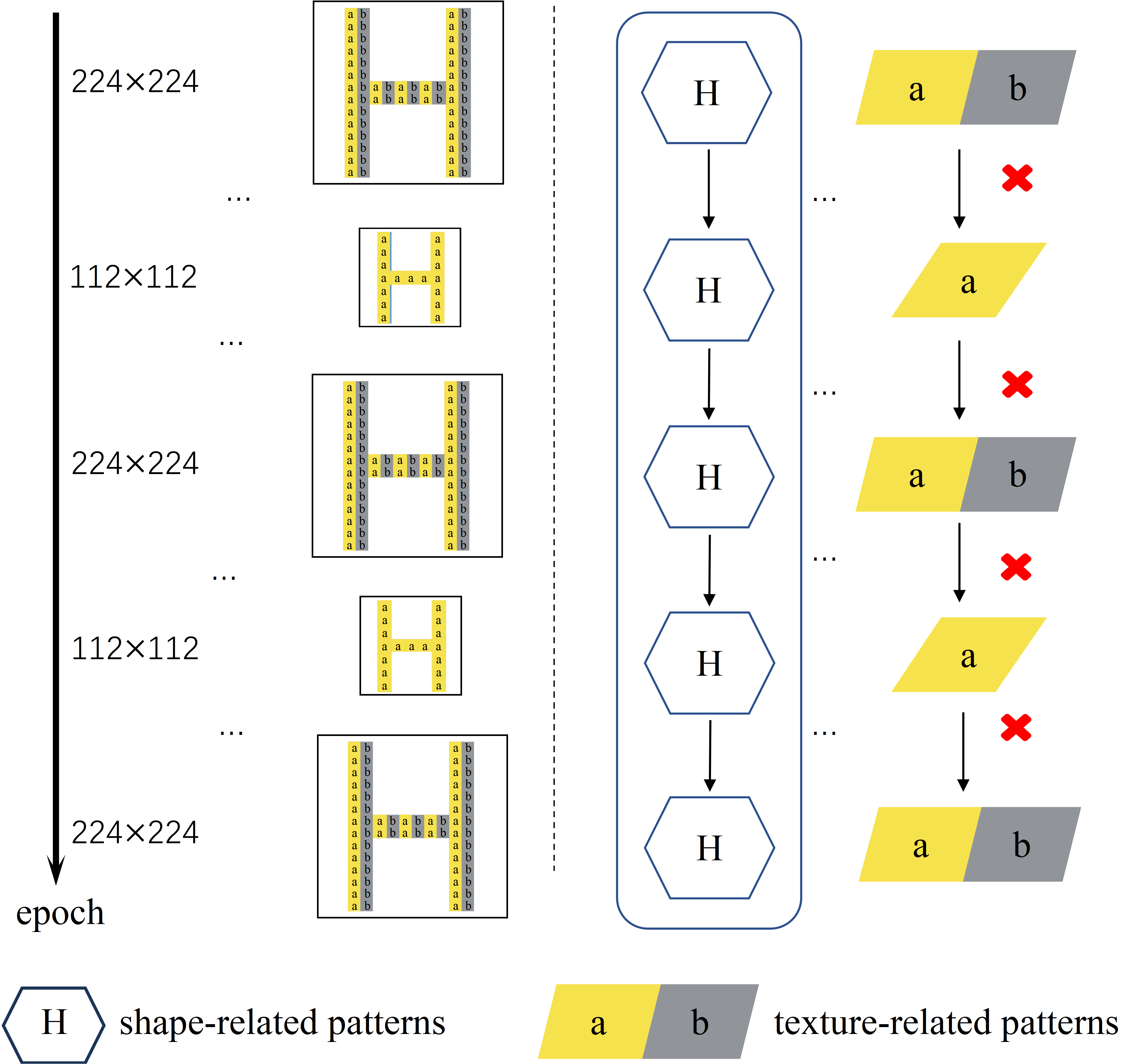} 
	\end{minipage}
	\hspace{0.01\linewidth}
	\caption{\textbf{Left:} The change of texture information and shape semantics when the image is reduced. The reduced image would lose large quantities of texture information (‘\textbf{a}’ and ‘\textbf{b}’), but the shape semantics (‘\textbf{H}’) still retains relatively completely. \textbf{Right:} Illustration of the proposed framework. The correlation between the texture-related patterns and the correct label becomes spurious since the texture-related patterns fail to hold in the same manner as they held in the past, enabling the model to focus on the more stable and invariant shape-related patterns.}
	\label{fig1}
\end{figure*}

In this paper, we focus on a novel perspective, \textit{consistency} of the shape semantics. As shown in the left part of Figure~\ref{fig1}, a large letter ‘\textbf{H}’ (shape) is rendered in small copies of some other lowercase letters ‘\textbf{a}’ and ‘\textbf{b}’ (texture). When we zoom out this image by simply taking one pixel every other pixel, the texture information would be left with only ‘\textbf{a}’, but, more importantly, the shape semantics retains relatively complete (we can still recognize ‘\textbf{H}’ through the reduced image). This provides a clear indication that compared with the changeability and fragility of the texture information, the shape semantics enjoys the consistency property that it remains stable during resolution change. We call this property \textit{shape consistency}. The above observation naturally leads to a question: since the shape pattern maintains complete semantics during resolution change, could temporally changing resolution during training help the model capture shape semantics? 

To answer this question, we first review existing resolution related methods. A series of works focus on adjusting image resolution~\cite{yang2020resolution, touvron2019fixing, wang2020resolution,  elad2019mix}. For example, to address the \textit{discrepancy} between the size of the objects during training and testing, Touvron et al.\cite{touvron2019fixing} employ different train and test resolutions via a simple parameter adaptation. Another line of research~\cite{wang2020glance, ming2021dynamic,  huang2018multiscale, Wang_2019_CVPR, Guo_2020_CVPR} stems from the \textit{redundacy} existing in the images, which believes that a large portion of ‘easy’ samples can already be classified correctly in low resolution or under small network. For example, DRNet~\cite{ming2021dynamic} proposes a resolution predictor to ensure each input image to be resized to its optimal resolution. Overall, although achieving promising performance, existing resolution related methods mainly depend on multi-resolution or proper resolution, the shape consistency from resolution change seems to be ignored. 

Based on the above intuition, we propose a novel training strategy as the answer of the above question: Temporally Resolution Decrement (TRD). The whole algorithm is surprisingly simple: we randomly select part of epochs to train the network with low resolution inputs. Experiments demonstrate the comparison between different resolutions in time domain could help the model integrate spatial shape semantics, and the introduction of reduced images enables improved computational efficiency. This is the first time that the temporal change of image resolution is associated with the shape bias of the model, which also motivates us to utilize shape consistency to improve model's shape bias by resolution change. Different from other methods of adaptive inference that either require modifications on model architectures or demand extra fine-tuning training, our method does not make any change to the network structure. The whole method can be implemented with a few lines of code and directly integrated into almost any CNN with ease.

%\added{We consider that two conditions need to be met for making full use of shape consistency by  changing resolutions: (i) the model needs to obtain the image knowledge of different resolutions, (ii) the model needs to repeatedly compare different resolution information. The proposed strategy ensures that the training procedure meets these two conditions by changing the resolution in epoch, i.e. in each training epoch, the model is fed with the same resolution images and compares different resolutions on different epochs.  An epoch training contains all the images, which met the first condition. }

It's worth noting that the proposed framework ensures the unity of input resolution in the time domain, i.e., in each training epoch, the model is fed with images of the same resolution. As shown in the right part of Figure~\ref{fig1}, each epoch of training can be viewed as one time of complete cognitive process of the model. During the iterative cognitive process, the changeability of the texture information undermines the correlation between the texture-related patterns and the correct label due to the resolution change between different epochs. Consequently, the model has to focus on the invariant shape-related patterns, allowing itself to generalize on two different input resolutions. Put another way, such epoch-wise operation gives rise to the reliance of the model on the robust shape-based representation. 

We evaluate the performance of our method on ImageNet with the vanilla ResNet-50~\cite{he2016deep} which typically serves as the default architecture in numerous studies. Surprisingly, when saving 77\% calculation overhead (randomly reducing the training image to 84$\times$84 within 90\% epochs), our method still improves ResNet-50 from 76.32\% to 77.26\%. When saving 43\% calculation overhead (randomly reducing the training images to 84$\times$84 within 50\% epochs), the accuracy can be further improved to 78.18\%. Superimposed with the strongest training methods of ResNet-50 on ImageNet, our method achieves 80.42\% top-1 accuracy with saving 37.5\% calculation overhead. To the best of our knowledge this is the highest ImageNet single-crop accuracy on ResNet-50 under 224$\times$224 without extra data or distillation.

Besides the lower computation cost as well as the improved performance, our method is still appealing in the generalization on varying image resolutions. The model trained with our method not only performs well on the original scale, but also shows remarkable improvements on the images with a wide range of resolutions for faster inference. It's important to note that although CNNs are not explicitly designed with the intention of handling varying image resolutions, various fields such as detection or self-driving cars where the objects would not appear in a fixed size may stand to benefit from models with such ability. On the other hand, part of the superiority of human vision system lies in the robustness against resolution change. Children learn about many things of the world through picture books, but they grow up without being confused by varying size of objects.

In summary, we make the following principle contributions in this paper:
\begin{itemize}
\item Inspired by the consistency of shape semantics during image zooming, we propose TRD, an effective yet simple plug-in method suitable for almost any CNN-based model.
\item Albeit simple, our method greatly improves training and inference efficiency. It breaks through the traditional computation-vs-accuracy trade-off view of deep learning that high computation cost is required to achieve a competitive performance, creating a win-win situation on both sides where lower computation cost and the improved accuracy can be simultaneously obtained without cautious balancing.
\item Our method manifests sweeping generalization on varying image resolutions, leaning close to the human vision system that would seldom be fooled by changes in image size.
\end{itemize}

\section{Related Works}

\noindent \textbf{Image Resolution:} Recently, an increasing amount of literature has focused on adjusting image resolution, an important factor which directly determines the computational costs and the performance of CNNs. Among them, much of the available literature~\cite{wang2020glance, ming2021dynamic,  huang2018multiscale, Wang_2019_CVPR} deals with the question of spatial redundancy, that not all regions in an image are task-relevant. Wang et al.~\cite{wang2020glance} propose to process a sequence of relatively small inputs which are selected from the original image with reinforcement learning instead of original image to perform efficient image classifications. Zhu et al.~\cite{ming2021dynamic} involve a resolution predictor that is explored and optimized jointly with the desired network into training to calculate the expected resolution for each input image. Yang et al.~\cite{yang2020resolution} propose RANet that is composed of sub-networks with different input resolutions, reducing computational cost by avoiding unnecessary computation on high-resolution when samples can be predicted in low-resolution. Huang et al.~\cite{huang2018multiscale} train a cascade of intermediate classifiers throughout the network which would be adaptively applied during test time to maintain coarse and fine level features.

%Touvron et al.~\cite{touvron2019fixing} propose a training strategy that employs different train and test resolutions, aiming to solve the discrepancy between the size of the objects seen by the classifier at train and test time. However, our method differs from it in two important aspects: (i) Motivation. FixRes focuses on discrepancy and hopes to match the training and testing data distributions, while our method inspires from shape consistency and focuses on helping the model learn the shape semantics. (ii) Design. FixRes operates models at much higher resolution at test time or much lower resolution at train time, thus relies on manual fine-tuning for adaptation and test-time augmentations. Our method only randomly reduces the training image resolution without any other operations required. Besides, FixRes is only applicable to the mismatch situation caused by the specific data augmentations (RandomResized Crop and Center Crop in the standard practice of training ImageNet), while our method is applicable to all types of datasets and data augmentation.

Howard et al.~\cite{Howard2018} propose progressive training that incrementally increase the input resolution for computational efficiency, but it usually comes with the cost of accuracy drop. Tan et al.~\cite{tan2021efficientnetv2} improve this strategy on EfficientNetV2 by adjusting the regularization strength according to input resolution. Note that the training with progressively increasing resolution could not help the model utilize shape consistency, because the comparison of the image resolution in adjacent epochs is not obvious, which is different from our method that improves shape bias by repeatedly comparing two resolutions for shape consistency. 

Elad et al.~\cite{elad2019mix} propose MixSize, a stochastic training regime where the image size as well as the batch size would be modified at each optimization step. Our method differs from it in the following two aspects: (i) MixSize involves multiple image sizes at each batch of training, while our method is an epoch-wise operation wherein the image size would remain the same at each epoch of training for utilizing shape consistency. (ii) Simplicity. MixSize may require several changes on the project, involving multiple changes on input resolutions, changes on batch size, changes on gradients and on batch-norm layers. Yet our method has shown to improve the computational efficiency with negligible modifications required, which could be implemented in a few lines of code.

\noindent \textbf{Texture bias and shape bias of CNNs:} Recent works~\cite{brendel2019approximating, geirhos2018imagenet, ballester2016performance, shi2020informative} argue that CNNs have the texture bias: CNNs classify images mainly according to their texture while shape information seems hard for CNNs to understand. A lot of works are proposed in the hope for improving the shape bias for CNNs. Geirhos et al.~\cite{geirhos2018imagenet} use Stylized-ImageNet to help the model learn the shape-based representation. Li et al.~ \cite{li2020shape} augments training data with images with conflicting shape and texture information for training a debiased model. These methods seek the balance between texture and shape preference all by consuming lots of extra computational overhead to generate transferred images or eliminating texture information, while our method can improve the shape bias of CNNs while reducing the computational cost.

\noindent \textbf{Model acceleration techniques:} There is a large volume of studies describing ways to speed up the network inference of deep networks. The most common direction is to design lightweight and efficient network architectures, such as MobileNets~\cite{howard2017mobilenets, sandler2018mobilenetv2, howard2019searching}, CondenseNet~\cite{Huang_2018_CVPR}, ShuffleNet~\cite{Zhang_2018_CVPR}. A number of recent works also focus on pruning redundant network connections~\cite{li2016pruning, liu2017learning, wang2021accelerate, luo2017thinet, tang2021manifold, Guo_2020_CVPR}. Since deep networks typically possess many redundant weights, some other works focus on quantizing the weights in the hope for faster inference~\cite{Jacob_2018_CVPR, hubara2016binarized}.

\section{Our Approach}
Temporally Resolution Decrement (TRD) is an effective albeit simple training strategy for CNNs. Instead of training the network with fixed image resolution, we randomly reduce the training image to low resolution in time domain. Specifically, each epoch has a $P$ probability of being selected into the set $\pi$, we train CNNs with low resolution images when present training epoch $e \in \pi$, and normally train with original resolution on the other epochs $E - \pi$, $i.e.$, the variable resolution $R$ of training images is set as:

$$ R=\left\{
\begin{array}{lcl}
R_{original}       &      & {e \in E - \pi}\\
\lambda     &      & {e \in \pi}\\
\end{array} \right. $$

where $e$ represents the present training epoch, and $R_{original}$ as well as $\lambda$ represent the original and reduced resolution, respectively. Note that we set epoch as the time slice rather than batch or even a single image for resolution change in time domain, because each epoch training contains whole training images and is able to ensure model obtain the present resolution knowledge. When the network changes from one resolution to   only the shape semantics is consistent while the texture information is difficult to guarantee uniformity due to its instability. Thus the correlation between the learned texture-related patterns and the correct label becomes spurious since the texture-related patterns fail to hold in the same manner as them held in the past. This would enforce the model to focus on the invariant and robust shape-related patterns, which has been proven to be beneficial in various ways. Our method enjoys a plug-and-play property that it can be conveniently incorporated into almost any existing deep learning project by merely adding a few lines of code. We present the code-level description of TRD algorithm in Appendix A.

\section{Experiments}
In this section, we investigate the effectiveness as well as the generalization of our method for several major computer vision tasks. We first evaluate the computational efficiency of TRD  with various architectures on ImageNet. Next, we study the performance of TRD on varying image resolution. Further, we explore why TRD works by designing different strategies. Besides, we test the generalization of our method to different datasets, networks, tasks and its compatibility with other regularization methods on image classification. The introduction of the experiment platform can be seen in Appendix B.

%All experiments are performed with Pytorch~\cite{paszke2017automatic}.  on Tesla M40 GPUs.

\subsection{TRD significantly improves computational efficiency on the ImageNet classification. }
ImageNet-1K~\cite{russakovsky2015imagenet} contains 1.2M training images and 50K validation images labeled with 1K categories. We first compare TRD with regularization methods on ImageNet with ResNet-50~\cite{he2016deep} using traditional training settings: all models are trained from scratch for 300 epochs with batch size 256 and the learning rate is decayed by the factor of 0.1 at epochs 75, 150, 225 for fair comparison. Then, we use state-of-the-art training procedure~\cite{wightman2021resnet} containing various data augmentation and regularization methods on 600 training epochs to test TRD's compatibility. Finally, we explore the performance of TRD on deeper network ResNet-101~\cite{he2016deep}. Detailed implementation settings of compared methods and training strategies are supplemented in Appendix B. FLOPs for network training are also evaluated. Our method uses different training resolutions on different epochs, so we calculate the weighted average value of training FLOPs in time domain as mFLOPs for fair comparison. We test the performance of our method with three reduced resolutions $\lambda$: 112, 84, 56 under three participation rates $P$: 0.5, 0.7, 0.9. The results are illustrated on Table~\ref{tab1}. 

%(Appendix)
%For Cutout~\cite{devries2017improved}, the mask size is set to 112$\times$112 and the location for dropping out is uniformly sampled. The hyper parameter $\alpha$ in Mixup~\cite{zhang2017mixup} is set to 1. (Appendix)
%We use the standard augmentation setting for ImageNet dataset including resizing, cropping, and flipping. 
%We evaluate classification accuracy on the validation set and the highest validation accuracy is reported over the full training course following the common practice.

\begin{table}[h]
	\begin{center}
		\caption{Summary of mFLOPs and validation accuracy of the ImageNet classification results based on various architectures.  ‘mFLOPs’ is weighted average value of  training FLOPs in time domain.  ‘*’ means results reported in the original papers.}
		%\renewcommand\tabcolsep{9pt} 
		%\vspace{1em} 
		\begin{tabular}{lccccc}
			\toprule[1.2pt]
			\midrule
			Method    &mFLOPs (G)   &Train   &Test     &Top-1 (\%)     &Top-5 (\%)        \\ \midrule
			ResNet-50(224$\times$224)                  &4.1   &224   &224    &76.32   &92.91      \\
			+Cutout~\cite{devries2017improved}         &4.1   &224   &224    &77.07   &93.32      \\
			+Mixup~\cite{zhang2017mixup}               &4.1   &224   &224    &77.42   &93.65      \\
			+AutoAugment*~\cite{cubuk2018autoaugment}  &-      &224   &224    &77.63   &93.82      \\
			+DropBlock*~\cite{ghiasi2018dropblock}     &-      &224   &224    &78.13   &94.02      \\
			\midrule
			+TRD ($P$=0.5)     &2.6 (\textcolor[RGB]{0,150,0}{-37\%$\uparrow$})    
			&112\&224   &224  &78.16   &94.04                         \\
			+TRD ($P$=0.7)     &2.0 (\textcolor[RGB]{0,150,0}{-52\%$\uparrow$})    
			&112\&224   &224  &77.96   &93.96                         \\
			+TRD ($P$=0.9)     &1.3 (\textcolor[RGB]{0,150,0}{-67\%$\uparrow$})       
			&112\&224   &224  &77.71   &93.84                         \\
			\midrule                           
			+TRD ($P$=0.5)      &2.4 (\textcolor[RGB]{0,150,0}{-43\%$\uparrow$})      
			&84\&224   &224   &78.18   &94.13                     \\
			+TRD ($P$=0.7)      &1.6 (\textcolor[RGB]{0,150,0}{-60\%$\uparrow$})       
			&84\&224   &224   &77.80   &93.91                     \\
			+TRD ($P$=0.9)      &0.9 (\textcolor[RGB]{0,150,0}{-77\%$\uparrow$})       
			&84\&224   &224   &77.26   &93.68                     \\
			\midrule                           
			+TRD ($P$=0.5)      &2.2 (\textcolor[RGB]{0,150,0}{-47\%$\uparrow$})      
			&56\&224  &224   &77.62   &93.82                     \\
			+TRD ($P$=0.7)      &1.4 (\textcolor[RGB]{0,150,0}{-66\%$\uparrow$})       
			&56\&224  &224   &76.75   &93.47                     \\
			+TRD ($P$=0.9)      &0.6 (\textcolor[RGB]{0,150,0}{-84\%$\uparrow$})       
			&56\&224  &224   &75.20   &92.57                     \\
			\midrule
			+Procedure D~\cite{wightman2021resnet}     &4.1                                            &224       &224     
			&80.01                 &94.75    \\
			+Procedure D \& TRD         &2.6 (\textcolor[RGB]{0,150,0}{-37\%$\uparrow$})     &112\&224   &224     
			&{\bfseries80.42}      &{\bfseries95.07 }   \\
			\midrule
			ResNet-101 (224$\times$224)               &7.8   &224        &224   
			&78.06            &94.12   \\   
			+TRD ($P$=0.5)             &4.9 (\textcolor[RGB]{0,150,0}{-37\%$\uparrow$})    &112\&224        &224    
			&79.34            &94.70    \\ 
			\midrule
			\bottomrule[1.2pt]
		\end{tabular}
		\label{tab1}
	\end{center}
\end{table}

\noindent \textbf{Higher network performance:}
We first measure the accuracy of the proposed method by comparing with data augmentation and regularization methods. Note that the improvement of TRD varies from 0.43\% to 1.86\% under different parameter settings. In particular, randomly reducing the training image resolution from 224$\times$224 to 84$\times$84 on 50\% epochs achieves 78.18\% top-1 accuracy on ResNet-50, which is better than the performance of ResNet-101 (78.06\%). Cutout~\cite{devries2017improved} and Mixup~\cite{zhang2017mixup} are representative data augmentation methods that generate more useful data from existing ones, and AutoAugment~\cite{cubuk2018autoaugment} uses reinforcement learning to find a combination of existing augmentation policies. However, TRD ($\lambda = 84, P= 0.5$) outperforms Cutout, Mixup and AutoAugment by +1.11\%, +0.76\% and +0.55\%, respectively.

Next, we verify whether the improvement of TRD can be continued in strong training strategy. Wightman et al.~\cite{wightman2021resnet} propose a series of competitive training procedures that integrate different optimization \& data augmentation methods. We choose Procedure D from these procedures including four data augmentation methods~\cite{cubuk2018autoaugment,zhong2020random,zhang2017mixup,yun2019cutmix} and three regularization methods~\cite{srivastava2014dropout,larsson2016fractalnet,szegedy2016rethinking} as a new baseline and combine TRD with this training strategy. Detailed implement of Procedure D is supplemented in Appendix B. Note that TRD can still improve the performance of ResNet-50 form 80.01\% to 80.42\% (+0.41\%) on such a strong baseline with reducing 37.5\% computational overhead, which further proves the effectiveness and the compatibility of our method. To the best of our knowledge this is the highest ImageNet single-crop accuracy of ResNet-50 at resolution 224$\times$224 without extra data or distillation. 

Besides, we notice that even if training images in 90\% epochs are reduced to low resolution 84$\times$84, TRD can still bring significant gain to the network performance (+0.94\%),  which shows that training with only 10\% normal images can still get better performance, as long as let the network learn the shape information by the comparison between different resolutions. Generally, training network only with low resolution images would cause drop in accuracy, but these results imply that low resolution images can help the network training by providing the comparison with original images in time domain. 

\noindent \textbf{Lower training computational cost:}
We also test average training FLOPs for TRD. The results in Table~\ref{tab1} show that our method can greatly reduce the training  overhead varying from 37\% to shocking 84\%, which is the attach benefit of shrinking the image resolution: the matrix and floating-point operation will be greatly reduced when the network is trained with low resolution images. Note that TRD ($\lambda=84, P=0.9$) can still improve 0.94\% higher accuracy than baseline with saving 77\% computing overhead, which proves that fixed resolution training would waste large quantities of computing resources especially on ImageNet. These results also illustrate that TRD can accelerate network training, dynamically save memory occupation in large cluster scenarios such as Google Cloud TPU, and thus reduce carbon emissions during training. Besides, the network performance declines only when reducing 90\% training images to 56$\times$56 (one sixteenth of the area for the original image), which shows the parameter selection of our method is very broad and TRD can create a win-win stituation on computation cost and network performance without cautious balancing. 

\begin{figure*}[h]
	\centering
	\hspace{0.01\linewidth}%
	\begin{minipage}[c]{0.32\textwidth} 
		\centering 
		\includegraphics[width=\linewidth]{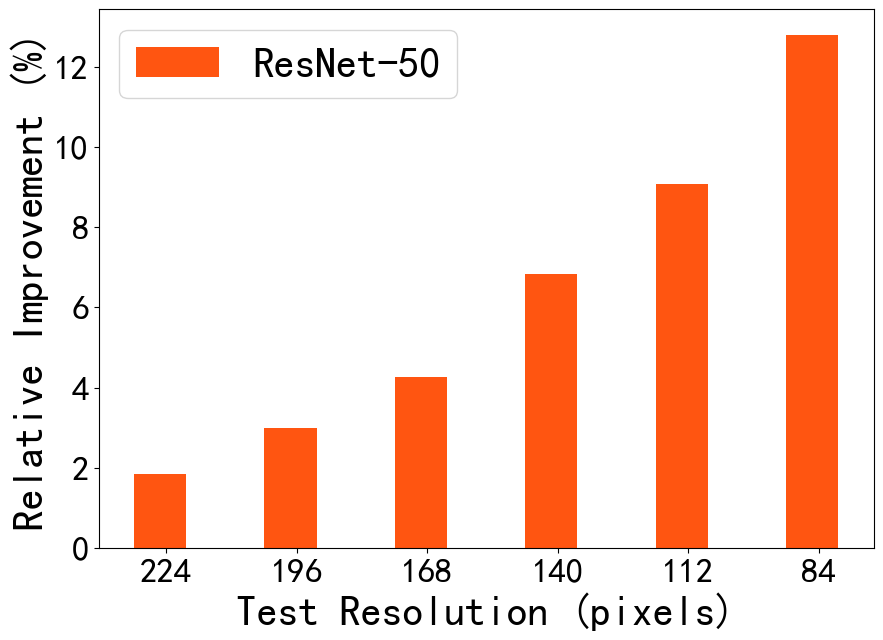} 
	\end{minipage}
	\hfill 
	\begin{minipage}[c]{0.32\textwidth} 
		\centering 
		\includegraphics[width=\linewidth]{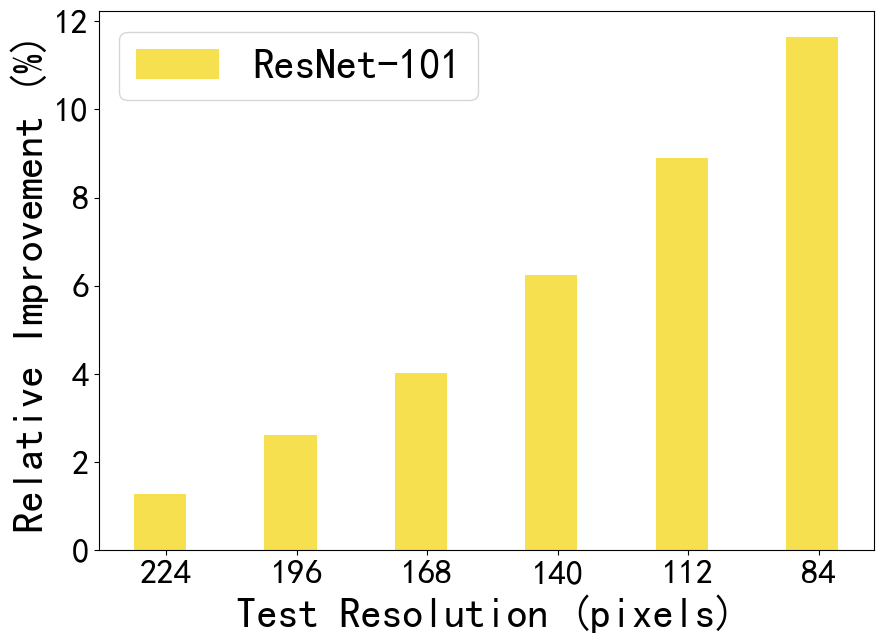} 
	\end{minipage}
	\hfill 
	\begin{minipage}[c]{0.32\textwidth} 
		\centering 
		\includegraphics[width=\linewidth]{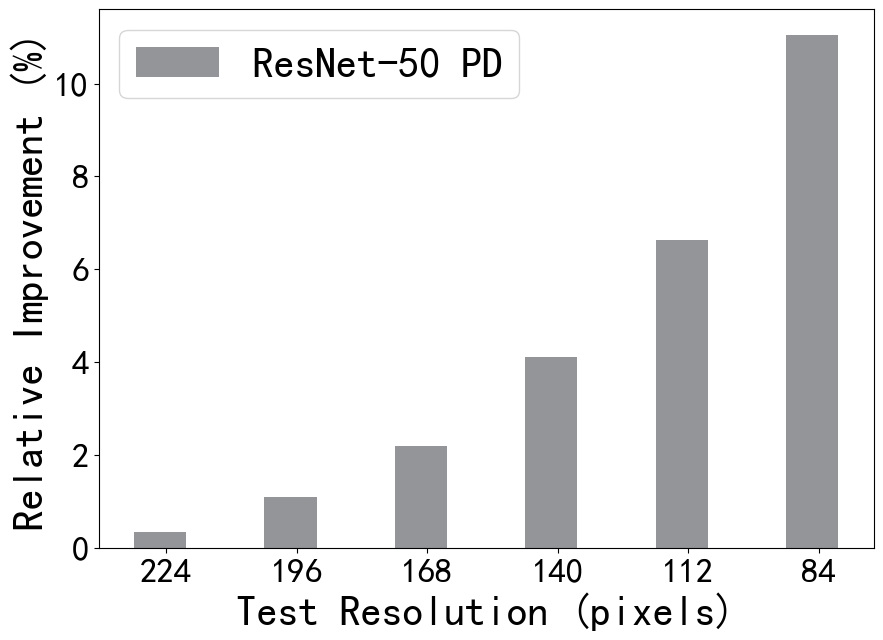} 
	\end{minipage}
	\hspace{0.01\linewidth}
	\caption{Relative improvement of TRD on different resolutions with ResNet-50, ResNet-101 and ResNet-50 PD (ResNet-50 trained with Procedure D). }
	\label{fig5}
\end{figure*}

\subsection{TRD achieves astonishing generalization on varying image resolutions.}
Besides standard testing where images are resized to 224$\times$224 on ImageNet, we also perform a stress test on validation set of ImageNet. We compare the performance of ResNet-50, ResNet-101 and ResNet-50 PD (ResNet-50 trained with Procedure D) with and without TRD ($\lambda=112, P=0.5$) on various inference resolutions from 224$\times$224 to 84$\times$84. Figure~\ref{fig5} shows the relative improvement of top-1 accuracy with TRD. TRD improves the generalization of all models on various resolutions, and this improvement expands with the widening gap between training resolution (224$\times$224) and inference resolution. The network trained with fixed resolution only remembers the texture information of its own scale, so its accuracy drops quickly on other scales. But TRD can improve model’s generalization on varying image resolutions by capturing more shape representations, so the higher the resolution difference, the higher the relative improvement of TRD. Although we train the network with only two resolutions, our method can improve the recognition accuracy for images with all resolutions, which also proves that our method makes the model learn the invariance among different resolutions, i.e., the shape-based representation, instead of simply remembering the texture information of two resolutions.

%\begin{figure}[h]
%\centering
%\includegraphics[width=0.8\columnwidth]{pic/pic5.png}
%\caption{Relative improvement of TRD on different test resolutions compared with baseline model.}
%\label{fig5}
%\vspace{-0.8em}
%\end{figure}

\subsection{TRD could also reduce the inference cost by simply reducing inference resolution.}

\begin{wraptable}{r}{9cm}
	\centering
	\caption{Comparison of inference efficiency on ImageNet.}
	\label{tab2}
	\begin{tabular}{lcc}
		\toprule[1.3pt]
		\midrule
		Models                            &Test FLOPs (G)       &Top-1 (\%)        \\
		\midrule
		ResNet-50 (224$\times$224)        &4.1            &76.32  \\
		\midrule
		SSS-ResNet-50~\cite{huang2018data}   &2.8 (-32\%)     &74.2   \\  
		Versatile-ResNet-50~\cite{wang2018learning}   &3.0 (-27\%)     &74.5   \\
		PFP-A-ResNet-50~\cite{liebenwein2019provable}   &3.7 (-10\%)     &75.9   \\
		C-SGD70-ResNet-50~\cite{ding2019centripetal}    &2.6 (-37\%)     &75.3   \\
		\midrule
		RANet~\cite{yang2020resolution}                 &2.3 (-44\%)     &74.0   \\ 
		DR-ResNet-50~\cite{ming2021dynamic}             &3.7 (-10\%)     &77.5   \\
		DR-ResNet-50 ($\alpha=2.0$)        &2.3 (-44\%)     &75.3   \\
		\midrule
		TRD-ResNet-50 (140$\times$140)    &1.6 (\textcolor[RGB]{0,150,0}{-60\%})    &74.77   \\
		TRD-ResNet-50 (168$\times$168)    &2.3 (\textcolor[RGB]{0,150,0}{-44\%})    &76.87   \\
		TRD-ResNet-50 (196$\times$196)    &3.2 (\textcolor[RGB]{0,150,0}{-23\%})    &{\bfseries78.71}  \\
		\midrule
		\bottomrule[1.3pt]
	\end{tabular}
\end{wraptable}

The above experiments demonstrate that the model trained with TRD still performs well on lower resolutions. This merit of TRD could be used to save the computational overhead of inference, i.e., using a smaller inference resolution to achieve good performance. We compare TRD with other methods on ImageNet to verify the superiority of the proposed method. The compared methods are composed of the representative model compression methods such as Sparse Structure Selection (SSS)~\cite{huang2018data}, Versatile Filters~\cite{wang2018learning}, PFP~\cite{liebenwein2019provable}, C-SGD~\cite{ding2019centripetal} and some latest adaptive inference methods including RANet~\cite{yang2020resolution} and DRNet~\cite{ming2021dynamic}. These methods reduce the inference cost by compressing the network structure or adaptively adjusting the reference resolution. We use ResNet-50  pretrained with Procedure D \& TRD as TRD-ResNet-50 and only reduce the inference resolution for comparing. As shown in Table~\ref{tab2}, TRD could achieve 74.77\% accuracy with reducing surprising 60\% inference cost, while gains 2.39\% accuracy boost with 23\% computation reduction compared to the original ResNet-50 on ImageNet. TRD achieves higher inference efficiency than other methods by simply reducing the test resolution.

\subsection{TRD can be universally applied to different networks, datasets, and computer vision tasks. }
We explore TRD's generalization for different networks, datasets, and computer vision tasks. First, we extensively evaluate TRD on CIFAR dataset~\cite{krizhevsky2009learning} with various architectures: VGGNet-19-BN ~\cite{simonyan2014very}, ResNet-56, ResNet-110, DenseNet-100-12~\cite{huang2017densely}, DenseNet-190-40 and Wide ResNet-28-10~\cite{zagoruyko2016wide}. Experiments show that our method consistently offers significant performance gains over the baselines of different structures. Second, We extend the experiments to various benchmark datasets. In particular, we use the following six benchmark datasets: CIFAR-10, CIFAR-100,Tiny-ImageNet, ImageNet, CUB-200-2011~\cite{wah2011caltech} and Stanford Dogs~\cite{khosla2011novel}. TRD is suitable for datasets with different image resolutions, various image quantity and unequal numbers of classification kinds, such as an impressive 4.8\% top-1 error reduction of ResNet-50 on Stanford Dogs. Finally, we show TRD can also be applied for training object detector in Pascal VOC~\cite{everingham2010pascal} dataset. We use RetinaNet~\cite{lin2017focal} framework as the baseline and compare the ResNet-50 backbone pretrained with and without TRD. The model pre-trained with TRD achieves higher accuracy (+1.47\%) than the baseline performance. Due to the space limitation, the detailed experimental settings and result analysis can be seen in Appendix C.

\subsection{Discussion} 
\label{limit}
\noindent \textbf{Network visualization:}  To analyze what the model learns with TRD, we visually compare the activation maps for our method against the baseline, using the corresponding class activation maps (CAM)~\cite{zhou2016learning} on two image resolutions shown in Figure~\ref{cam}. More CAM visualization examples can be seen in Appendix D. 

\begin{figure*}[h]
	\centering
	\includegraphics[width=0.9\linewidth]{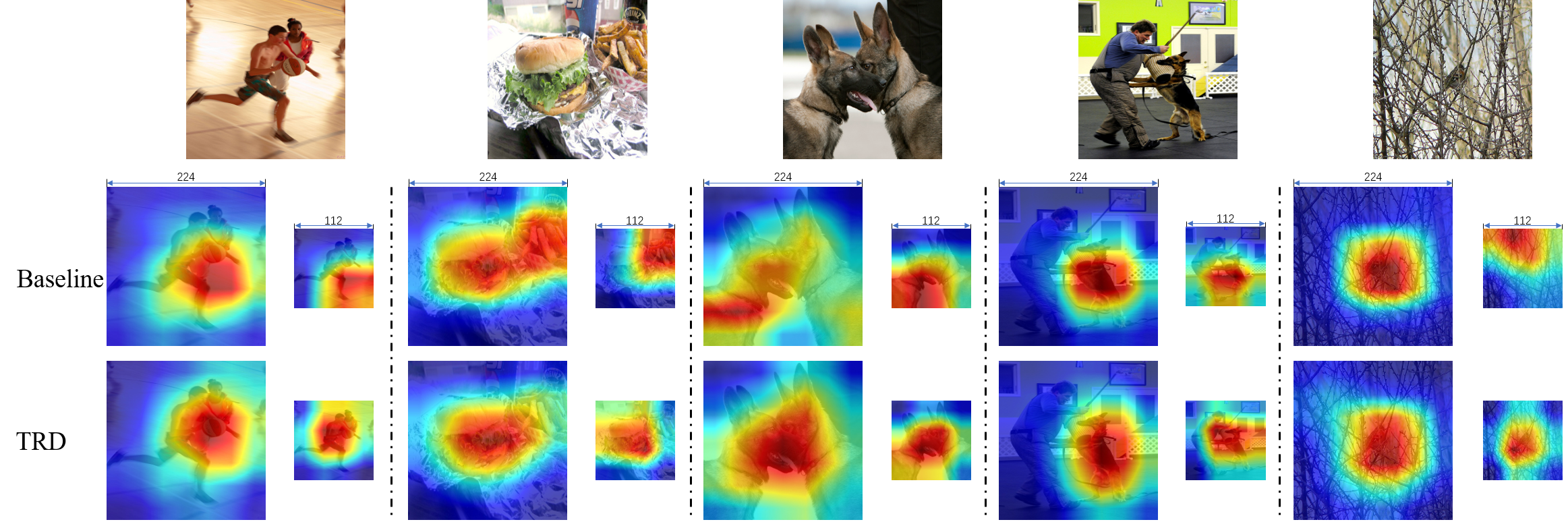}
	\caption{Class activation mapping (CAM)~\cite{zhou2016learning} visualizations on samples with two image resolutions (224$\times$224 and 112$\times$112).  ‘Baseline’ denotes the vanilla ResNet-50 model to clearly see the effect of our method. Note that TRD can capture targets more accurately, and shows a relatively uniform perception pattern under different image resolutions.}
	\label{cam}
\end{figure*}

We notice that TRD successfully forces the model to understand the content of the image from the perspective of shape. Consequently, the model trained with TRD captures the targets more accurately. For example, when the cheeseburger is placed right next to the French fries, the baseline model would take the fries into consideration, which is in fact a symptom of shortcut learning~\cite{geirhos2020shortcut}. On the contrary, the model trained with TRD that absorbs more shape semantics handles this co-occurrence situation pretty well. Another interesting observation with the figure is that the model trained with TRD maintains a relatively uniform perception of images with varying resolutions, whilst the perception of baseline model would shift seriously when the testing image resolution mismatches with the original training resolution. These observations not only demonstrate the superiority of the perception of shape semantics, but also explain why TRD can help the model generalize on varying image resolutions.

\noindent \textbf{Why dose TRD work? Resolution comparison v.s Multi-resolution:} We explore why simply reducing resolution in time domain could help the model capture more shape semantics and improve the computational efficiency. Due to the simplicity of TRD, we consider the the gain may come from two aspects: the comparison between different resolutions and/or the learning of multi-resolution. We verify this suppose by designing different training strategies as shown in Table~\ref{tab6}. 

\begin{wraptable}{r}{8cm}
	\centering
	\caption{Performance of TRD variants on Tiny ImageNet.}
	\label{tab6}
	\begin{tabular}{p{5cm}rp{1.5cm}}
		\toprule[1.2pt]
		\midrule
		Methods                                   
		&Top-1 (\%)                     \\
		\midrule
		ResNet-110 (64$\times$64)                     &62.42                      \\
		+Proposed TRD ($\lambda=32, P=0.5$)    
		&{\bfseries 64.57}        \\
		\midrule
		+Three-Resolutions TRD                &64.26            \\
		+Mix-Resolutions TRD                  &64.01            \\
		+Batch-Wise TRD                       &61.32            \\
		+Regular TRD                          &64.21            \\
		\midrule
		\bottomrule[1.2pt]
	\end{tabular}
\end{wraptable}

(i) ‘Three-Resolutions TRD’ randomly reduces training images in half epochs to 32$\times$32 or 48$\times$48. ‘Mixed-Resolutions TRD’ randomly reduces training images in half epochs to images ranging from 32$\times$32 to 64$\times$64. Both strategies containing more resolutions information than our TRD with only two, but their performances are lower than TRD. This observation clearly proves that multi-resolution in time domain does not bring performance improvement, it is the comparison between different resolutions that actually helps network capture more shape semantics due to shape consistency. 
(ii) ‘Batch-Wise TRD’ randomly reduces 50\% training images into 32$\times$32 in every training batch, but it reduces the network performance by 1.1\%. The premise of the comparison between images with different resolutions is that the network fully grasps the image information of the current resolution. Each batch contains different images, thus it is difficult to make the network learn the different resolutions of the same image in batch iteration. It also demonstrate that epoch is the optimal choice for resolution change in time domain. 
(iii) ‘Regular TRD’ reduces training images in each even number of epoch to 32$\times$32, which achieves similar performance to random strategy. It proves that the order of resolution is not important, and the comparison between resolution is the core for TRD.

\noindent \textbf{Limitation:} 	Through empirical results, we have demonstrated the significant boosts in both computational efficiency and generalization on varying image resolutions that  TRD provides in different network structures, data sets and computer visual tasks. However, the generalization of these results is subject to certain limitations. The effectiveness of TRD relies on an underlying premise that a shape-based decision rule is the intended or at least a useful solution. However in some special application scenarios such as medical imaging, a model needs to utilize local texture details to perceive visual stimuli meaningfully. Such an additional uncontrolled factor may lead to the instability of the positive effect TRD could bring on training as it has done in other general computer vision fields. The exact effect of TRD on medical imaging will be explored in the near future.

\section{Conclusion}
In this paper, we observe that the texture information would be devastated while the shape semantics maintains a relatively complete form during the image zooming operation. This consistency of shape semantics drives us propose a new training strategy: TRD. We randomly select some training epochs with a certain probability and reduce the training images to a smaller resolution in these epochs. It can be seen that TRD is very simple and requires mere several lines of code modification. Training with different resolutions in epoch-wise mode makes the network capture the shape semantics of training images, and surprisingly improves the computational efficiency. The robust shape-based representation also pushes forward to generalization on varying image resolutions, which conforms to superiority of the human vision system.

\section{Broader Impact}
Image resolution is an important factor in training CNNs, which directly relates to the accuracy as well as the computation cost of neural networks. It is therefore crucial to study the effect of image resolution on network training. We view the variable resolution method presented in this paper as an important step towards this goal. Inspired by the consistency of shape semantics during image zooming, we associate the change of resolution in time domain with the enhancement of shape semantics for the first time, which is ignored by existing resolution related methods. Besides, we empirically demonstrate that the comparison between different resolutions is the key to capture shape semantics, and epoch is the optimal time slice for resolution change. These findings would be benefit for the research community because it provides a new approach to improve the shape bias of CNNs from the perspective of resolution change. Moreover, some actual scenes like automated driving put higher requirements for both prediction accuracy and response time, but it is difficult to give consideration for both better performance and faster training or inference. TRD breaks the deadlock by improving accuracy, reducing training computation cost and accelerating inference with a few lines of code. It can also be superimposed with state-of-the-art methods and universally applied to various computer vision tasks. Our method conforms to the conception of Green AI~\cite{schwartz2020green} that appeals for pushing forward the development of AI with fewer carbon emissions. Therefore, we advocate incorporating TRD into the paradigm of CNNs' training.

{\small
	\bibliographystyle{ieee_fullname}
	\bibliography{re}
}

\appendix

\section{Code-level Description of TRD Algorithm}

We present the code-level description of TRD algorithm in Algorithm~\ref{al}. We set a random seed evenly distributed between 0 and 1. At the beginning of each epoch, if the value of random seed is less than the threshold, i.e., the participation\_rate $P$, the training images of each batch in this epoch would be reduced to a specific smaller size, i.e., the reduced resolution $\lambda$. Note that in practice we usually do not reduce image resolution in the two epochs before and after the learning rate decreases when adopting a step-decay policy, to avoid the probabilistic training instability. It can be seen that TRD is very simple and requires mere several lines of code modification, so it can be easily used in different tasks for higher computational efficiency. 

\begin{algorithm}[h]
	\caption{Pseudo-code of TRD}
	\label{al}
	\LinesNumbered
	\ForEach{epoch}{
		$flag = 0$ \\
		\If{random.random() $<$ Participation\_rate}{$flag=1$  \hfill  \# randomly select some training epochs}
		\ForEach{iteration}{
			input, target = get\_minibatch(dataset)
			
			\If{mode == training}{
				\If{flag==1}{input = F.interpolate(input, size=[$\lambda$, $\lambda$]) \hfill \# reduce images to a smaller resolution}
				output = model\_forward(input)   
				
				loss = compute\_loss(output, target)
				
				model\_update()
			}
		}
	}
\end{algorithm}

\section{Implementation Details}
We describe the training implementation settings in detail. All experiments are performed with Pytorch~\cite{paszke2017automatic} on Tesla M40 GPUs except Procedure D related experiments that are performed on NVIDIA GeForce RTX 3080Ti.  

We first compare TRD with regularization methods on ImageNet with ResNet-50 using traditional training settings: all models are trained from scratch for 300 epochs with batch size 256 and the learning rate beginning from 0.1 is decayed by the factor of 0.1 at epochs 75, 150, 225 for fair comparison. The standard augmentation setting including resizing, cropping, and flipping is used for ImageNet dataset. We evaluate classification accuracy on the validation set and the highest validation accuracy is reported over the full training course following the common practice. For Cutout~\cite{devries2017improved}, the mask size is set to 112$\times$112 and the location for dropping out is uniformly sampled. The hyper parameter $\alpha$ in Mixup~\cite{zhang2017mixup} is set to 1.

Next, we verify whether the improvement of TRD can be continued in strong training strategy. Wightman et al.~\cite{wightman2021resnet} propose a series of competitive training procedures that integrate different optimization \& data augmentation methods. We choose Procedure D from these procedures as a strong baseline. In Procedure D, the ResNet-50 is trained on 600 epochs with batch size of 384, AdamP is chosen as the optimizer with a cosine learning rate decay and binary cross-entropy. Initial learning rate is set to 0.0033 and wight decay is set to 0.01. Label Smoothing~\cite{szegedy2016rethinking} (P=0.1), Dropout~\cite{srivastava2014dropout} (P=0.1), and Stochastic Depth~\cite{larsson2016fractalnet} (P=0.05) are chosen as the regularization methods, RandAugment~\cite{cubuk2018autoaugment} (M=7, N=3, MSTD=1.0), Mixup~\cite{zhang2017mixup} ($\alpha=0.2$), CutMix~\cite{yun2019cutmix} ($\alpha=1.0$), and Random Erasing~\cite{zhong2020random} (Prob=0.35, Count =1) are chosen as the data augmentation methods. 

\section{Generalization of TRD for different architectures, datasets, and computer vision tasks}

\subsection{TRD can be universally applied to various architectures.}
We extensively evaluate TRD on various architectures: VGGNet-19-BN ~\cite{simonyan2014very}, ResNet-56, ResNet-100, DenseNet-100-12~\cite{huang2017densely}, DenseNet-190-40 and WideResNet28-10~\cite{zagoruyko2016wide}. All the experiments were performed on CIFAR-10 and CIFAR-100~\cite{krizhevsky2009learning}. For fair comparison, all the experiments strictly use the same preprocessing and data augmentation strategies such as random flipping and cropping, following the common practice. For TRD, we test the performance with randomly reducing the training image resolution from $32\times32$ to $16\times16$ on 30\% and 50\% of epochs. Notably, the results in Table~\ref{tab1} show that our method consistently offers significant performance gains over the baselines on different structures, such as an impressive 1.37\% top-1 error reduction on ResNet-110. It also means our method can reduce the running time of large-scale network and make the training of small network faster on the basis of improving the accuracy. Besides, the resolution of images in CIFAR-10 and CIFAR-100 is 32$\times$32, the network can still gain from 16$\times$16 images training, which proves TRD's generalization for various image resolutions.

\begin{table*}[h]
	\begin{center}
		\begin{tabular}{p{1.5cm}p{2.5cm}p{2.6cm}p{2.6cm}p{2.6cm}}
			\toprule[1.2pt]
			\midrule
			Datasets
			&Models
			&\begin{tabular}[c]{@{}l@{}}Vanilla\\ \footnotesize{Top-1 / mFLOPs} \end{tabular}
			&\begin{tabular}[c]{@{}l@{}}TRD ($P$=0.3)\\ \footnotesize{Top-1 / mFLOPs} \end{tabular}
			&\begin{tabular}[c]{@{}l@{}}TRD ($P$=0.5)\\ \footnotesize{Top-1 / mFLOPs} \end{tabular}        \\
			\midrule
			\multirow{6}{*}{CIFAR-10}&ResNet-56        &93.84\% / 90.3M                        &94.76\% / 70.0M                        &94.37\% / 56.5M             \\
			&ResNet-110       &94.32\% / 173.3M                        &95.32\% / 134.3M                       &95.07\% / 108.3M            \\
			&VGGNet-19-BN     &93.55\% / 399.5M                       &94.28\% / 309.6M                       &93.94\% / 249.7M            \\
			&DenseNet-100-12     &95.30\% / 297.9M                       &95.73\% / 230.8M                       &95.65\% / 186.2M            \\
			&DenseNet-190-40  &96.62\% / 9.4G                         &96.93\% / 7.3G                         &96.84\% / 5.9G              \\
			&Wide ResNet-28      &96.16\% / 5.3G                         &96.40\% / 4.1G                         &96.22\% / 3.3G              \\
			\midrule
			\multirow{6}{*}{CIFAR-100}&ResNet-56       &73.71\% / 90.3M                        &75.10\% / 70.0M                        &74.57\% / 56.5M             \\
			&ResNet-110       &74.71\% / 173.3M                        &76.54\% / 134.3M                      &75.70\% / 108.3M            \\
			&VGGNet-19-BN     &73.14\% / 399.5M                       &73.52\% / 309.6M                       &72.67\% / 249.7M            \\
			&DenseNet-100-12     &77.25\% / 297.9M                       &78.22\% / 230.8M                       &78.11\% / 186.2M            \\
			&DenseNet-190-40  &82.43\% / 9.4G                         &84.06\% / 7.3G                         &83.77\% / 5.9G              \\
			&Wide ResNet-28      &81.27\% / 5.3G                         &81.45\% / 4.1G                         &80.06\% / 3.3G              \\
			
			\midrule
			\bottomrule[1.2pt]
		\end{tabular}
	\end{center}
	\caption{Top-1 accuracy (\%) and mFLOPs with and without TRD using different architectures on CIFAR-10 and CIFAR-100. The results show that TRD is suitable for CNN models with different structures.}
	\label{tab1}
\end{table*}

\subsection{TRD is generally effective on multiple benchmark datasets.}	
We then extend the experiments to other benchmark datasets. In particular, we use the following six benchmark datasets: CIFAR-10 ($32\times32$), CIFAR-100 ($32\times32$),Tiny-ImageNet ($64\times64$), ImageNet ($224\times224$), CUB-200-2011~\cite{wah2011caltech} ($448\times448$) and Stanford Dogs~\cite{khosla2011novel} ($224\times224$). We consider these datasets should be able to cover a wide range of scenarios in the computer vision field. For TRD, we test the performance with randomly reducing the training image resolution to one quarter on 30\% and 50\% of epochs. As show in Table~\ref{tab5}, TRD is suitable for datasets with different image resolutions, various image quantity and unequal numbers of classification kinds. Besides, CUB-200-2011 and Stanford Dogs are the representive datasets of fine-grained image classification, the performance of TRD on these two datasets shows that the shape information still has a high gain effect even on fine-grained image classification which requires more subtle differences.
\begin{table*}[h]
	\begin{center}
		\begin{tabular}{p{2.2cm}p{2cm}p{2.5cm}p{2.5cm}p{2.5cm}}
			\toprule[1.2pt]
			\midrule
			Datasets
			&Models
			&\begin{tabular}[c]{@{}l@{}}Vanilla\\ \footnotesize{Top-1 / mFLOPs} \end{tabular}
			&\begin{tabular}[c]{@{}l@{}}TRD ($P$=0.3)\\ \footnotesize{Top-1 / mFLOPs} \end{tabular}
			&\begin{tabular}[c]{@{}l@{}}TRD ($P$=0.5)\\ \footnotesize{Top-1 / mFLOPs} \end{tabular}        \\
			\midrule
			CIFAR-10         &ResNet-110     &94.32\% / 173.3M      &95.32\% / 134.3M             &95.07\% / 108.3M               \\
			CIFAR-100        &ResNet-110     &74.71\% / 173.3M     &76.54\% / 134.3M             &75.70\% / 108.3M              \\
			Tiny ImageNet     &ResNet-110     &62.42\% / 693.3M    &64.39\% / 537.3M             &64.57\% / 433.3M              \\
			ImageNet          &ResNet-50      &76.32\% / 4.1G      &77.94\% / 3.2G               &78.16\% / 2.6G                \\
			CUB-200-2011      &ResNet-50      &66.73\% / 16.5G    &66.92\% / 12.5G              &66.54\% / 10.3G              \\
			Stanford Dogs     &ResNet-50      &61.38\% / 4.1G     &66.18\% / 3.2G               &63.12\% / 2.6G                \\
			\midrule
			\bottomrule[1.2pt]
		\end{tabular}
	\end{center}
	\caption{Top-1 accuracy (\%) and mFLOPs with and without TRD using different datasets on ResNet-110 and ResNet-50. The results show that TRD is suitable for different kinds of datasets.}
	\label{tab5}
\end{table*}

\subsection{TRD enables the improved object detection performance.}
In this subsection, we show TRD can also be applied for training object detector in Pascal VOC~\cite{everingham2010pascal} dataset. We use RetinaNet~\cite{lin2017focal} framework as the baseline and compare the ResNet-50 backbone pretrained with and without TRD. Models pretrained on ImageNet are fine-tuned on Pascal VOC 2007 and 2012 trainval data and evaluated on VOC 2007 test data using the mAP metric. We follow the fine-tuning strategy of the original method.

\begin{table}[h]
	\begin{center}
		\begin{tabular}{p{5.5cm}rp{1.5cm}}
			\toprule[1.2pt]
			\midrule
			Models                                   
			&mAP (\%)                     \\
			\midrule
			RetinaNet (baseline)                     &70.14                      \\
			+TRD Pre-trained ($\lambda$=112, $P$=0.5)        
			&{\bfseries 71.61}                        \\
			+TRD Pre-trained ($\lambda$=112, $P$=0.7)          &71.59                      \\
			\midrule
			\bottomrule[1.2pt]
		\end{tabular}
	\end{center}
	\caption{Object detection results on Pascal VOC with RetinaNet.}
	\label{tab3}
\end{table}

As shown in Table \ref{tab3}, the models pre-trained with TRD achieve higher accuracy than the baseline performance. The results suggest that the model trained with TRD can better capture the target objects.  Besides, although the models trained with TRD on different participation rates $P$ have different accuracy in image classification, their performance in object detection tends to be consistent. This finding offers a compellingly simple explanation for how does TRD help network training: TRD helps the network to learn the shape-based representation.

\section{More Cam Visualizations}
\begin{figure}[h]
	\centering
	\includegraphics[width=1\linewidth]{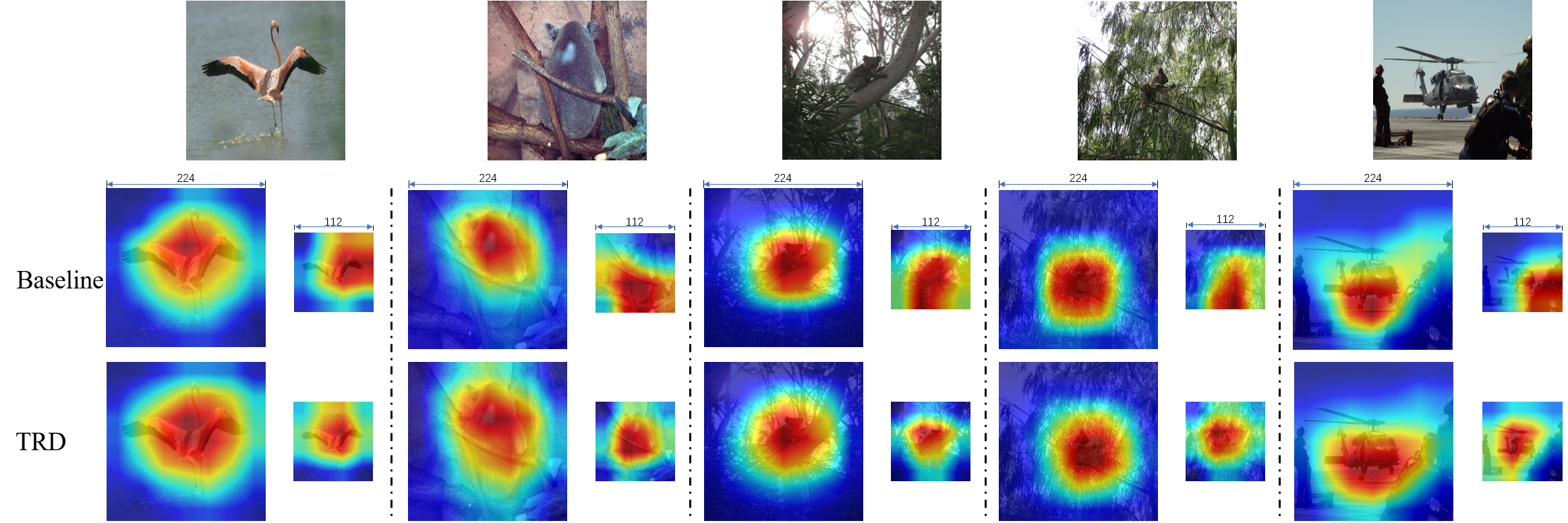}
	\caption{ Supplementary class activation mapping (CAM) visualizations on samples with two image resolutions (224 and 112). ‘Baseline’ denotes the vanilla ResNet-50 model to clearly see the effect of our method.}.
	\label{fig3}
\end{figure}

For better comparison, we include more cam visualization examples here in Figure~\ref{fig3}.

\section{Ablation Studies}
We conduct ablation studies on Tiny ImageNet dataset (64$\times$64) with ResNet-110.

\begin{figure}[h]
	\centering
	\includegraphics[width=0.95\columnwidth]{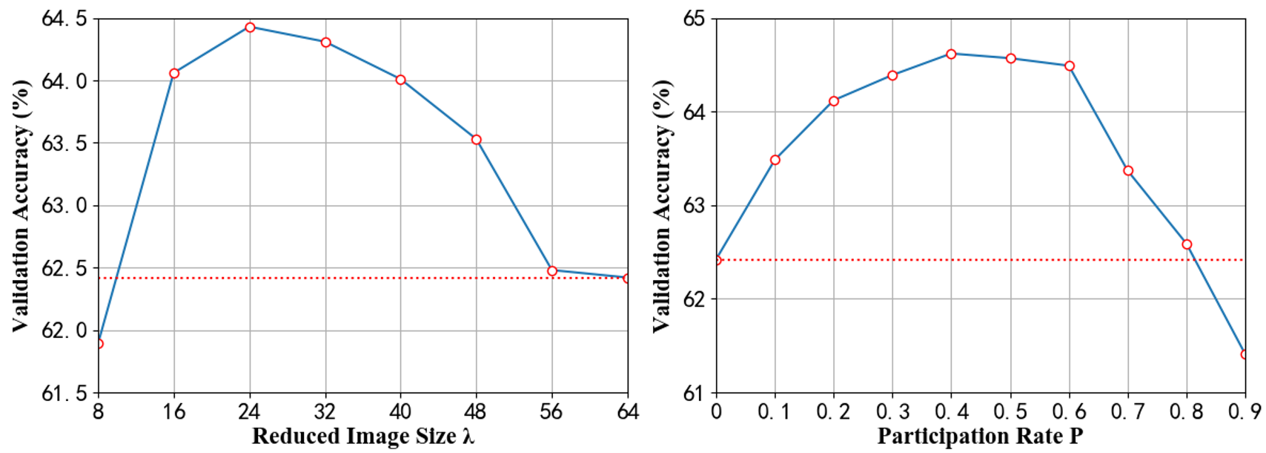}
	\caption{Tiny ImageNet validation accuracy of TRD on ResNet-110 against reduced image size ($\lambda$) and participation rate ($P$). Red dotted line indicates the baseline performance. }
	\label{fig6}
\end{figure}

\noindent \textbf{Analysis on reduced resolution \bm{$\lambda$}:} We evaluate TRD when reduced resolution $\lambda$ $\in$ \{8, 16, 24, 32, 40, 48, 56, 64\} with setting participation rate $P$ to 0.5 as shown in left part of Figure~\ref{fig6}. When $\lambda$ is set to 56, the performance improvement is not obvious. It is because the difference between 56$\times$56 image and 64$\times$64 image is not obvious, the network can not capture the invariance (shape reprensentation) from the comparison between two similar resolutions. With the continuous reduction of image resolution, network performance continues to improve. Note that the performance improvement (+1.68\%) can still be achieved when the image is reduced to one sixteenth (16$\times$16), which indicates that shape semantics can be preserved in very small images, and the choices for  parameter $\lambda$ are very broad. However, when $\lambda$ is set to 8, the performance of network is lower than baseline. The reason is that 8$\times$8 image is too small to carry basic shape semantics, which results in invalid training.

\noindent \textbf{Effect of the participation rate \bm{$P$}:} Specially, we call the ratio of epochs using reduced images to all epochs as participation rate $P$. We set the reduced resolution to 32$\times$32 and show the effect of the $P$ in right part of Figure~\ref{fig6}. With the increase of $P$, the accuracy first rises and then decreases after reaching the highest when the $P$ is 0.4, which means the network is trained with 60\% normal images and 40\% reduced images. Besides, reducing 80\% training images can still achieve the performance of baseline, but the amount of computation is saved by 60\%. When training with 90\% reduced images, the performance of the network decreases, which shows that only training with 10\% normal images can not make network learn enough information on Tiny ImageNet.

\end{document}